\documentclass[journal]{IEEEtran}
\usepackage{graphicx}
\usepackage{amsmath}
\usepackage{amssymb}
\usepackage{algorithm}
\usepackage{algorithmic}
\usepackage{multirow}
\usepackage{color}
\usepackage{booktabs}
\usepackage{array}\newcolumntype{C}[1]{>{\centering\let\newline\\\arraybackslash\hspace{0pt}}m{#1}}
\usepackage{epsfig}
\usepackage{graphicx}
\usepackage{amsmath}
\usepackage{amssymb}
\usepackage{amsfonts}
\usepackage{CJK}
\usepackage{amsmath}
\usepackage{algorithm}
\usepackage{algorithmic}
\usepackage{bm}
\usepackage{indentfirst}
\usepackage{multirow}
\usepackage{array}
\usepackage{mathrsfs}
\usepackage{color}
\usepackage{url}
\usepackage{subfigure}

\usepackage[marginal]{footmisc}
\usepackage{type1cm}
\usepackage{times}

\usepackage[colorlinks,
            linkcolor=blue]{hyperref}

\begin{document}
%
\title{Dual-Stream Reciprocal Disentanglement Learning for Domain Adaptation Person Re-Identification}
%
%
%

\author{Huafeng Li,
        Kaixiong Xu,
        Jinxing Li,
        Guangming Lu,
        Yong Xu,~\IEEEmembership{Senior Member,~IEEE},
        Zhengtao Yu,\\
        and~David Zhang,~\IEEEmembership{Life Fellow,~IEEE}


\thanks{The work is supported by the NSFC fund (61966021, 61906162), Shenzhen Science and Technology Program (RCBS20200714114910193), Open Project Fund from Shenzhen Institute of Artificial Intelligence and Robotics for Society (AC01202005017), and China Postdoctoral Science Foundation (2020TQ0319, 2020M682034).}

\thanks{H. Li, K. Xu, and Z. Yu are with the Kunming University of Science and Technology, Kunming, China. E-mail: hfchina99@163.com,  xukaixiong@stu.kust.edu.cn, ztyu@hotmail.com}
\thanks{J. Li (Corresponding Author), G. Lu  and Yong Xu are with Harbin Institute of Technology, Shenzhen, China, also with Shenzhen Key Laboratory of Visual Object Detection and Recognition, Shenzhen, China. E-mail: lijinxing158@gmail.com, luguangm@hit.edu.cn, yongxu@ymail.com}
\thanks{D. Zhang are with the Chinese University of Hong Kong (Shenzhen), also with the Shenzhen Research Institute of Big Data, Shenzhen, China. E-mail: davidzhang@cuhk.edu.cn}}

%
%

\markboth{Journal of \LaTeX\ Class Files}%
{Shell \MakeLowercase{\textit{et al.}}: Bare Demo of IEEEtran.cls for IEEE Journals}
%



\maketitle

\begin{abstract}
Since human-labeled samples are free for the target set, unsupervised person re-identification (Re-ID) has attracted much attention in recent years, by additionally exploiting the source set. However, due to the differences on camera styles, illumination and backgrounds, there exists a large gap between source domain and target domain, introducing a great challenge on cross-domain matching. To tackle this problem, in this paper we propose a novel method named Dual-stream Reciprocal Disentanglement Learning (DRDL), which is quite efficient in learning domain-invariant features. In DRDL, two encoders are first constructed for id-related and id-unrelated feature extractions, which are respectively measured by their associated classifiers. Furthermore, followed by an adversarial learning strategy, both streams reciprocally and positively effect each other, so that the id-related features and id-unrelated features are completely disentangled from a given image, allowing the encoder to be powerful enough to obtain the discriminative but domain-invariant features. In contrast to existing approaches, our proposed method is free from image generation, which not only reduces the computational complexity remarkably, but also removes redundant information from id-related features. Extensive experiments substantiate the superiority of our proposed method compared with the state-of-the-arts. The source code has been released in {\url{https://github.com/lhf12278/DRDL}}.%
\end{abstract}


%
\IEEEpeerreviewmaketitle

\section{Introduction}
\IEEEPARstart{P}{erson} re-identification (Re-ID) \cite{zheng2012reidentification,koestinger2012large, liao2015person, yang2014salient, 8640834, 8873614, 7999218}, \cite{Vishwakarma8469037, Vishwakarmajaiswal2019state, Vishwakarmayadav2020person} aims to match probe pedestrian images under various but non-overlapped cameras. Thanks to the rapid development of deep learning \cite{li2020drpl,li2019relaxed,li2017shared,li2018probabilistic} and the wide application prospects in other computer vision tasks such as action recognition and gesture recognition \cite{Vishwakarma8662635, VishwakarmaIntegratedapproach, Vishwakarma8960517, VISHWAKARMA2019157, Vishwakarma7279962},  the performances of person Re-ID have been remarkably improved in recent years. 
However, this achievement highly follows the hard constraint: both training and testing sets are collected from the same domain. It has been widely proved that applying a well trained Re-ID model to practical scenarios, it will meet a large performance degradation due to the domain gap between the training and testing images. Fortunately, unsupervised domain adaptation (UDA)\cite{Zhang8370105, Xu8946732, Peng8902166} person Re-ID provides a possible and efficient solution to address this problem.

\begin{figure}
\begin{center}
   \includegraphics[width=1\linewidth]{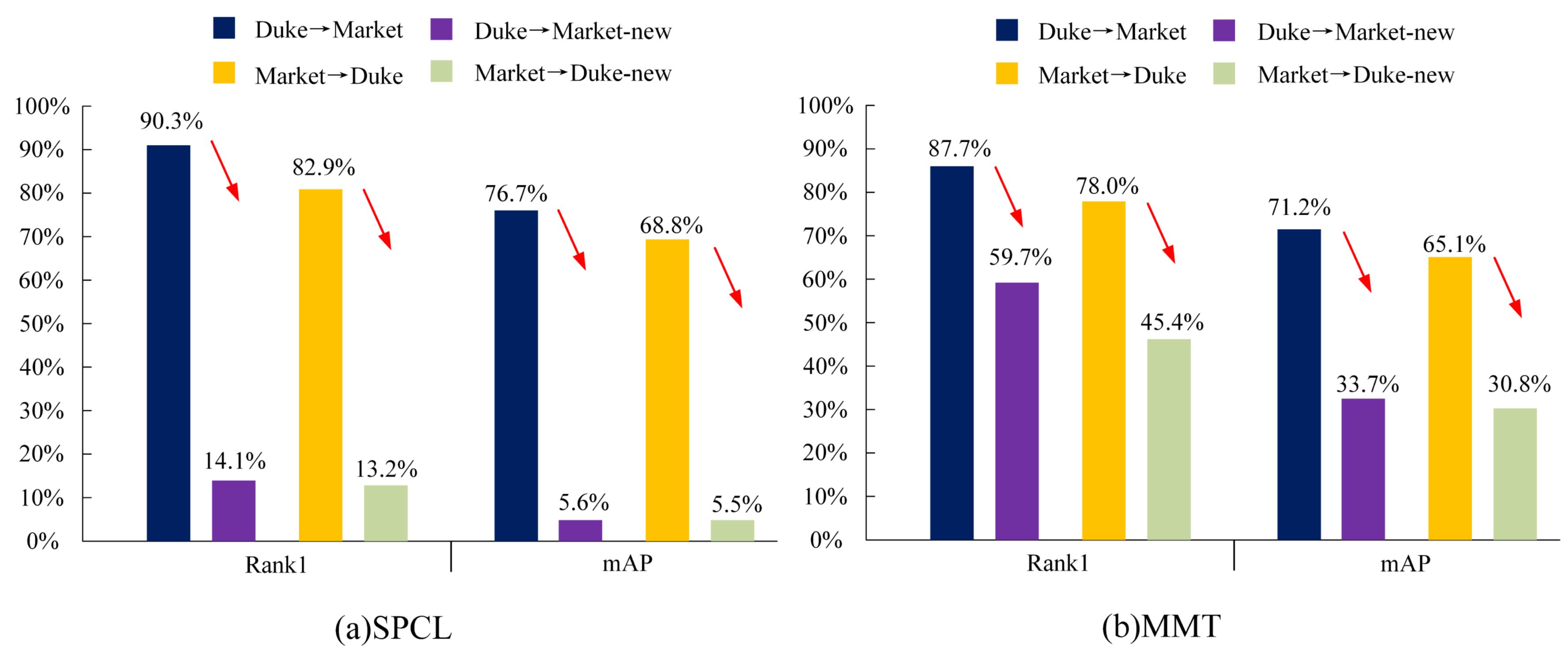}
\end{center}
   \caption{The experimental results obtained by SPCL \cite{ge2020self} and MMT \cite{ge2020mutual} on Duke$\rightarrow$Market, Market$\rightarrow$Duke, Duke$\rightarrow$Market-new, and Market$\rightarrow$Duke-new.}
\label{fig:bar}
\end{figure}

Existing UDA person Re-ID methods can be roughly divided into three categories: extra auxiliary model based methods \cite{wei2018person, kumar2020unsupervised, ren2019domain, yang2019leveraging, huang2019sbsgan, liu2019adaptive}, clustering (pseudo-label prediction) based methods \cite{yu2017cross, fan2018unsupervised, yu2018unsupervised, fu2019self, song2020unsupervised, ge2020mutual}, and domain-invariant feature learning based methods \cite{zhong2019invariance, yang2019patch, wu2019unsupervised, chang2019disjoint, zhang2019fuzzy, li2020structure}. Referring to extra auxiliary model based methods, they usually introduce an additional model to pre-process inputs, e.g., PT-GAN \cite{wei2018person}, ECN \cite{zhong2019invariance}, and SPGAN \cite{8578208}, subsequently removing the domain discrepancy from the source and target domains. However, due to this auxiliary model, it would heavily increase the computational complexity, decreasing the efficiency on practical applications.

For the clustering based methods, despite the fact that they achieve satisfactory performance on some datasets, they heavily follow an assumption: each instance is captured by enough cameras and has numbers of positive samples. Applying this kind of methods to the scenarios where there is a high possibility that a person only appears in a camera, the performance is inferiorly influenced. For instance, in the benchmark datasets  Market1501 (Market) \cite{zheng2015scalable} and DukeMTMC-reID (Duke) \cite{ristani2016performance}, each person was captured by at least two cameras. As shown in Fig.\ref{fig:bar}, following the setting in SPCL \cite{ge2020self} and MMT \cite{ge2020mutual}, satisfactory performances on them are achieved. However, as mentioned above, the case that a person appears in only one camera is quite general in a practical application, and the assumption in Market and Duke is not followed. To more reasonably meet the practical application, we re-conduct these two datasets, so that parts of identities were only attained by a camera (more detailed settings can be found in section \ref{section:practical}). Here we denote these re-conducted datasets as Market-new and Duke-new. From Fig.\ref{fig:bar} we can easily observe that both two state-of-the-art methods meet a great performance degradation, being far from our requirement. By contrast, although domain-invariant feature learning based methods have obtained slightly lower performance on Re-ID since there is no pseudo-label or auxiliary model, they are free from the aforementioned limitations and easily applied to various real-world applications.  Note that, we do not argue that clustering based Re-ID methods are influenced by the number of cameras, but are their number of positive samples. The key constraint of clustering is the number of positive samples belonging to the same identity. In other words, we should make sure that each identity has numbers of samples captured from various cameras. However, in practical applications, this constraint is too strict. Thus, in this paper, we also mainly focus on them.


\begin{figure}
\begin{center}
   \includegraphics[width=1\linewidth]{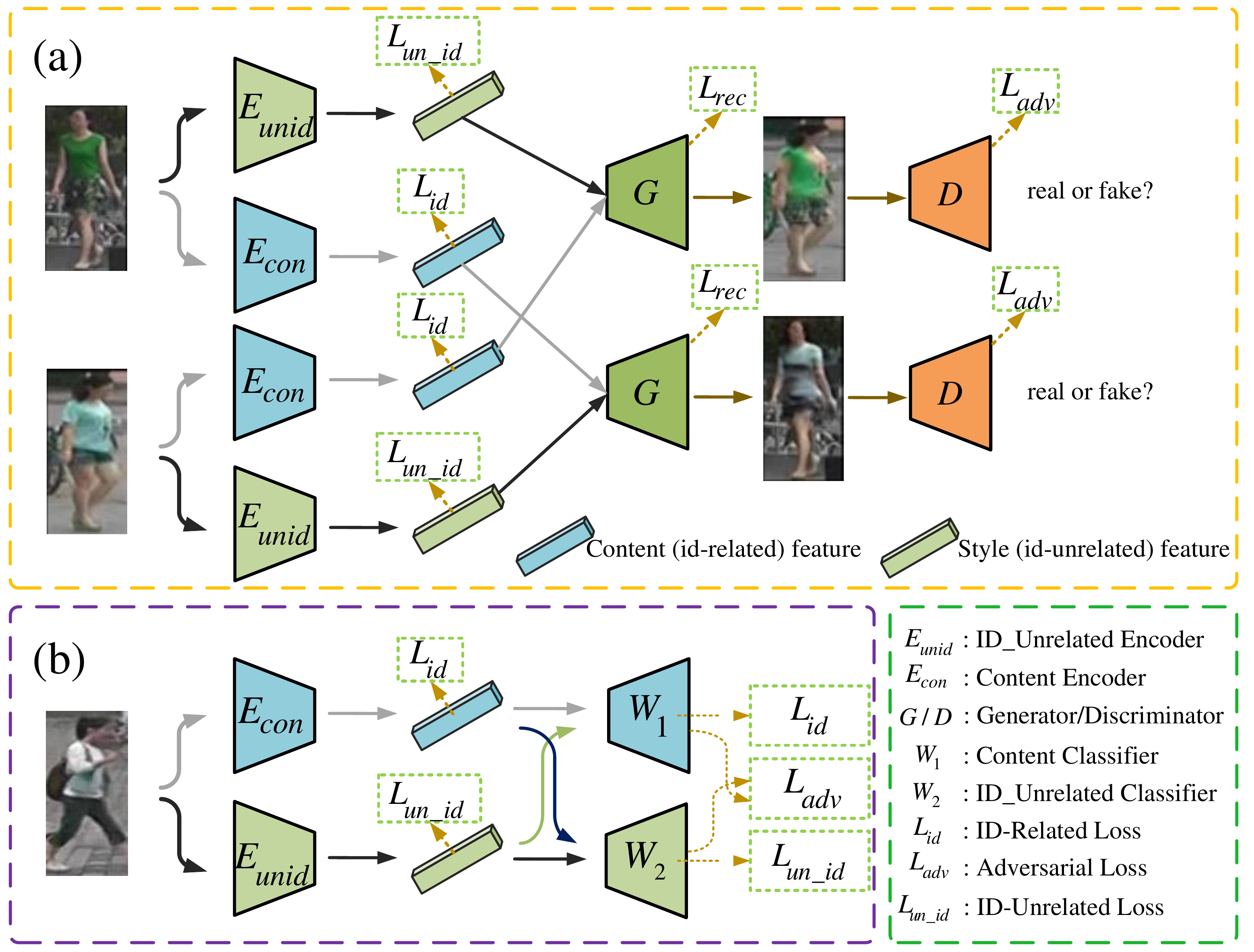}
\end{center}
   \caption{The frameworks of disentanglement learning. (a) existing methods\cite{zheng2019joint,zou2020joint,liu2018exploring,huang2018multimodal,li2019cross,choi2020hi}; (b) our proposed method. Note that, existing disentanglement learning methods are usually constraint to the image generation, which may subsequently introduce redundant or id-unrelated information. By contrast, our proposed method is free from image generation and mainly focuses on learning id-related but domain-unrelated transformations.}
\label{fig:examp}
\end{figure}



Referring to existing domain-invariant feature learning methods, disentanglement learning is a general but efficient technique for person Re-ID. Typical examples are shown in Fig.\ref{fig:examp}(a) \cite{zheng2019joint,zou2020joint,liu2018exploring,huang2018multimodal,li2019cross,choi2020hi},  which obtain both id-related and id-unrelated features. Specifically, by combining these two types of features to reconstruct images, the id-related features enjoy better semantic information. \textbf{Despite these approaches achieved a remarkable improvement in Re-ID, they are highly dependent on the qualities of the generated images. Due to this constraint, redundant information, like color, illumination or backgrounds etc., also must be extracted exactly. However, it is really intractable that a network is capable of extracting all of these information, subsequently resulting in an inferior model learning.} Although the spectrum-disentangled representation learning (SDL) \cite{kansal2020sdl} is proposed to avoid image reconstruction, it also encourages the sum of id-related and id-unrelated features to recover the feature associated with the input image by using the L2 loss. Furthermore, in SDL, a classifier is exploited to enforce the id-unrelated feature to be averagely classified to each identity. However, for the id-related feature, there is no any measurement between it and id-unrelated information. In other words, it is possible that the id-related feature is corrupted by some id-unrelated information.


\textbf{In fact, it has been proven that, to achieve the pedestrians matching, it is unnecessary to extract the whole person-related features from an image, but only discriminative features associated with the identity are enough \cite{meng2020mutual}}. Inspired by this, in this paper a novel dual-stream reciprocal disentanglement learning (DRDL) is proposed to simply but efficiently learn domain-invariant features for unsupervised person Re-ID. As shown in Fig.\ref{fig:examp}(b), two streams are constructed for id-related feature learning and id-unrelated feature learning, respectively. The former one is used for extracting semantic features corresponding to different persons, and the latter one is used for extracting domain features corresponding to different cameras. Particularly, the id-related features and id-unrelated features are directly extracted from the given images. By following an adversarial strategy, the id-unrelated features are exploited to enforce the id-related features to be more robust, and these learned domain-invariant features are then inversely beneficial for the id-unrelated feature extraction. \textbf{Note that, compared with existing methods, our proposed method is completely free from the image generation or the feature reconstruction, efficiently preventing from introducing redundant information caused by generated images with poor qualities.}

The main contributions of this paper contain:
\begin{itemize}
\item
Instead of domain-crossly generating novel images, our proposed method constructs a two-stream network to directly learn the id-related and id-unrelated features, avoiding introducing redundant information.

\item
By following an adversarial strategy, an id-related (content) classifier and an id-unrelated classifier are embedded to reciprocally disentangle id-related/id-unrelated features, greatly allowing us to achieve domain-invariant feature extraction.

\item
The presented approach is much simpler and more efficient, which also obtains better performance on several person Re-ID datasets compared with state-of-the-arts.
\end{itemize}

The rest of this paper is organized as follows. In Section \ref{section:related_works}, some related works about unsupervised Re-ID methods are briefly described. The proposed approach DRDL is then introduced in Section \ref{section:proposed_method}. To demonstrate the effectiveness of DRDL, we conduct experiments on several datasets in Section \ref{section:experiments}, followed by the conclusion in Section \ref{section:conclusion}.
\section{Related Works}\label{section:related_works}

In this section, we briefly introduced various UDA person Re-ID methods, including extra auxiliary model based methods, clustering based methods, and domain-invariant feature learning based methods.

\subsection{Extra Auxiliary Model based Methods}

The extra auxiliary model based methods for person Re-ID are to exploit additional models to transfer the image styles or change the human body poses. Based on these generated as well as raw images, the domain gap is removed. Here the Generative Adversarial Network (GAN) \cite{goodfellow2014generative} based methods are the most general ones. Specifically, these methods focus on transferring the labeled images from a source domain to a target domain, encouraging the styles of transferred samples to be consistent to that in the target domain, so that the domain discrepancy caused by various cameras are removed. Inspired by this analysis, PT-GAN \cite{wei2018person}, SPGAN \cite{8578208}, ATNet \cite{liu2019adaptive} and CR-GAN \cite{chen2019instance} were proposed. However, these methods ignore the variations (illumination and color, etc.) in the target domain, which limit the Re-ID performance enhancement. Thus, Zhong et al. \cite{zhong2019invariance} proposed three constraints: instance-invariant, camera-invariant and domain-invariant, which efficiently solve the problems mentioned above. Additionally, considering the changes on body poses, Li et al. \cite{li2019cross} primarily extracted the pose associated information and then introduced GAN to learn pose-invariant and domain-invariant features. Although these extra auxiliary model based methods can obtain additional images for supervised learning, they are too sensitive to qualities of the generated images.

\subsection{Clustering Based Methods}
The clustering based methods aim to use the clustering algorithms to predict pseudo-label or soft-label for each trained sample in the target domain, so that the model can be supervised trained. In order to remove the noise-label from the estimated labels, clustered samples are then filtered through an asymmetric co-teacher strategy, in which different teacher networks are reciprocally updated \cite{yang2020asymmetric}. Similarly, Zhao et al. \cite{zhao2020unsupervised} presented a robust training strategy which also introduces two networks to reciprocally assign confidential samples. Instead of designing dual networks, according to the similarity, the self-similarity grouping method \cite{fu2019self} was proposed to adaptively group samples for each view, and then these groups are labeled for supervised learning. Since there exist partial samples which are quite difficult to be classified and heavily affect the similarity measurement, Zeng et al. \cite{zeng2020hierarchical} simultaneously introduced the hierarchical clustering technique and embedded hard-triplet loss to reduce their inferior influence. Except for the dual-network and similarity based methods, the mutual mean-teaching (MMT) \cite{ge2020mutual} provides another technique to make label prediction. Specifically, this approach firstly gets partial but the confidential labels through an offline strategy. Based on these samples with confidential labels, remaining samples are further labeled by exploiting an online algorithm. Experimentally, if each identity is captured in several cameras and has enough positive images, these clustering based pseudo-label prediction methods are capable of achieving satisfactory performance on Re-ID. However, in many practical scenarios, there is a high possible for a person appears in only one camera, greatly limiting their applications.

\subsection{Domain-Invariant/Disentanglement Learning based Methods}

Different from aforementioned approaches, domain-invariant feature learning based methods do not predict any pseudo-label or use extra models to generate images, but only aim to learn domain-invariant features. For instance, a novel network was presented to transfer features from a labeled source set to an unlabeled target set \cite{wang2018transferable}. Considering the variance caused by distinctive cameras, Wu et al. \cite{wu2019unsupervised} proposed a Camera-Aware Similarity Consistency Loss to learn consistent pairwise similarity distributions for intra-camera matching and cross-camera matching. To remove the id-unrelated features, Zou et al. \cite{zou2020joint} jointly disentangled id-related/unrelated features and encouraged adaptation to work on the id-related feature space exclusively. Furthermore, \cite{zheng2019joint} successfully separated the features into the structure part and body part, which are cross-combined for data argumentation, improve the robustness of the model on different scenarios. Since videos contains more space-time information, Zhao et al. \cite{zhao2019attribute} proposed an attribute-driven approach for feature disentanglement learning.  Kansal et al.\cite{kansal2020sdl} replaced image reconstruction with feature reconstruction to realize the disentanglement of id-related and id-unrelated features for visible-infrared person re-ID.

Although various methods have been proposed, it is still challenged to efficiently separate the id-related features and the id-unrelated features. The main reason is that existing methods fall into crossly combining these two types of features of different samples to make reconstruction, which may not only increase the difficulty of person identification for networks, but also introduce redundant information. To tackle this issue, we propose a novel disentanglement learning method, which is quite simple but achieves much better performance.

\begin{figure*}
\begin{center}
   \includegraphics[width=0.95\linewidth]{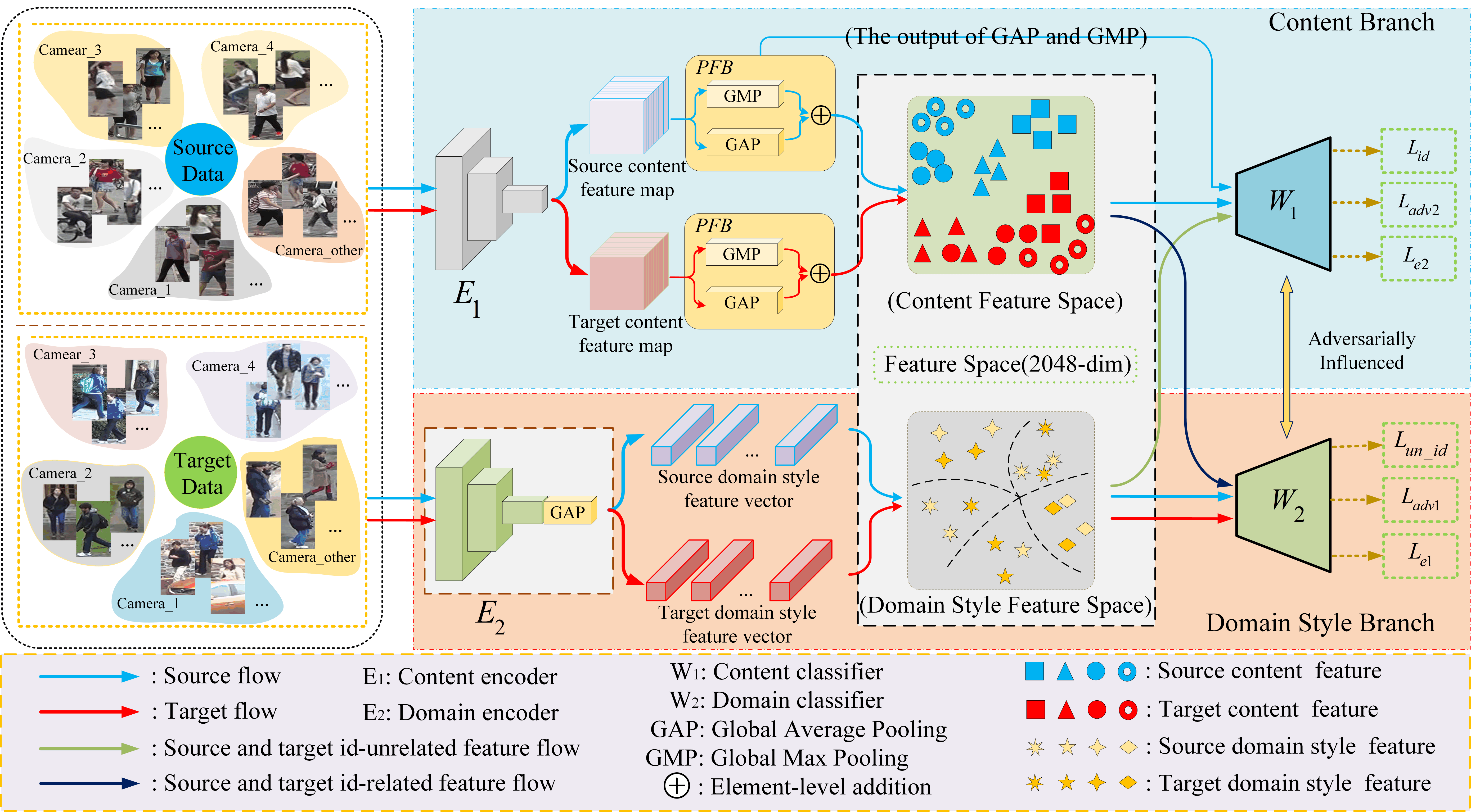}
\end{center}
   \caption{The pipeline of our proposed method. It consists of two streams, where the first stream is used for domain-invariant feature extraction and the second stream is used of domain-style feature extraction.}
\label{fig:pipeline}
\end{figure*}

\section{Proposed Method}\label{section:proposed_method}
The pipeline of the proposed method is shown in Fig.\ref{fig:pipeline}. As we can see, there are two streams named content branch and domain style branch, where each one consists of an encoder and a classifier. Specifically, given the source and target images, the content encoder $E_{1}$ is exploited for id-related but domain-invariant feature extraction, which is measured in a supervised way by its corresponding classifier $W_{1}$ based on the assignments of the source dataset. By contrast, in the second stream, the domain encoder $E_{2}$ is designed for id-unrelated feature extraction, which is also handled by the classifier $W_{2}$. Furthermore, as depicted at right in Fig.\ref{fig:pipeline}, each stream reciprocally and positively influences another stream to further update the encoders and classifiers. Thanks for this strategy, the domain-invariant features and id-unrelated features are respectively extracted.

Technically, the ResNet50 \cite{he2016deep} pretrained on the ImageNet \cite{deng2009imagenet} is regarded as the backbone of encoders. Max pooling preserves the most salient features, while other information is ignored. In contrast, average pooling combines all information in a weighted way, while the salient features are limited. Therefore, max pooling and average pooling are complementary to each other and their combination is beneficial for more meaningful feature learning. Here we use both GAP and GMP at the end of $E_{1}$, and feed the output of $E_{1}$ through Pooling Fushion Block (PFB) to get the final representation. Additionally, since the id-unrelated features are relatively easy to get, we directly apply the GAP to $E_{2}$. For classifiers, $W_{1}$ consists of BN (batch normalization) and FC (fully-connected layer); $W_{2}$ is composed of a Convolution layer (for the feature map size reduction: from 2048-D to 32-D), BN, ReLu and FC.


\subsection{Model Pretrain}

To make $(E_{1}, W_{1})$ and $(E_{2}, W_{2})$ be capable of classifying id-related features and id-unrelated features, we first conduct pretraining for these two streams. Denote $\mathbf{X}^{s}=\{\mathbf{x}_{i}^{s}\}_{i=1}^{N^{s}}$ as the source domain set, where $\mathbf{x}_{i}^{s}$ is its $i$-th instance and $N^{s}$ is its total number. Specifically, $\mathbf{x}_{i}^{s}$ is associated with a person ID $y_{i}^{s}\in\{1,2,\cdots, M^{s}\}$ and a camera ID $c_{i}^{s} \in \{1,2,\cdots, C^{s}\}$, where $M^{s}$ and $C^{s}$ are the numbers of identities and cameras, respectively. Similarly, for the target domain, denote $\mathbf{X}^{t}=\{\mathbf{x}_{j}^{t}\}_{j=1}^{N^{t}}$ as the target set, and each sample $\mathbf{x}_{j}^{t}$ is corresponding to a camera ID $c_{j}^{t}\in\{C^{s}+1, C^{s}+2, \cdots, C^{s}+C^{t}\}$, where $N^{t}$ and $C^{t}$ are the numbers of samples and cameras in the target domain, respectively. The output of id-related features of the source set and the target set through $E_{1}$ is represented as $f\_content_{i}^{s}$ and $f\_content_{j}^{t}$, respectively. Note that, the person IDs are not available for the target set.

Since $E_{1}$ and $W_{1}$ are used for id-related feature extraction and classification, the source set is introduced to update them in a supervised way based on existing person IDs. As described above, to obtain more discriminative features, the final id-related features can be represented as
\begin{equation}
\begin{aligned}
f\_content_{i}^{s}=GAP(E_{1}(\mathbf{x}_{i}^{s}))+GMP(E_{1}(\mathbf{x}_{i}^{s}))
\end{aligned}\label{eq:E1}
\end{equation}
where $GAP(\cdot)$ and $GMP(\cdot)$ denote global average pooling and global max pooling, respectively. Through inputting aforementioned three features into $W_{1}$, the outputs and given person IDs can be measured by using the cross-entropy loss.
\begin{equation}
\begin{aligned}
L_{id}(E_{1},W_{1})=
&CE\left(W_{1}\left(GAP(E_{1}(\mathbf{x}_{i}^{s}))\right),y_{i}^{s}\right)\\
&+CE\left(W_{1}\left(GMP(E_{1}(\mathbf{x}_{i}^{s}))\right),y_{i}^{s}\right)\\
&+CE\left(W_{1}\left(f\_content_{i}^{s}\right),y_{i}^{s}\right)
\end{aligned}\label{eq:Lpersonid}
\end{equation}
where $CE(\cdot)$ denotes the cross-entropy loss.

Additionally, to enforce $E_{2}$ and $W_{2}$ to be capable of extracting the id-unrelated features and classifying them, the camera assignments for both source domain and target domain are exploited for the measurement.
\begin{equation}
\begin{aligned}
&L_{un\_id}(E_{2},W_{2})\\=
&CE\left(W_{2}\left(E_{2}(\mathbf{x}_{i}^{s})\right),c_{i}^{s}\right)+
CE\left(W_{2}\left(E_{2}(\mathbf{x}_{j}^{t})\right),c_{j}^{t}\right)
\end{aligned}\label{eq:Lcameraid}
\end{equation}
Note that, images obtained by different cameras meet different domains, and each domain contains its corresponding id-unrelated information, e.g., color, illumination, and backgrounds, etc. Thus, it is reasonable to use the camera IDs to encourage $E_{2}$ to extract the id-unrelated information.

Thanks for the pretraining according to Eq.(\ref{eq:Lpersonid}) and Eq.(\ref{eq:Lcameraid}), the two encoders enjoy the powerful capabilities of extracting id-related and id-unrelated features, respectively. However, $E_{1}$ only focuses on the source domain, while given an unseen testing data whose domain is possibly different from the source domain, the learned model would meet a great performance degradation. Thus, we further introduce the reciprocal disentanglement learning to enforce the id-related features to be domain-invariant.

\subsection{Dual-Stream Reciprocal Disentanglement Learning}

To achieve the domain-invariant feature learning, here encoders and classifiers from the two streams are crossly combined. On the one hand, the content encoder $E_{1}$ and the camera classifier $W_{2}$ are jointly taken into account to remove any domain-related information from the learned id-related features; on the other hand, the domain encoder $E_{2}$ and the id classifier $W_{1}$ are then trained to encourage $E_{2}$ to remove any id-related information. Following these two steps, the id-related and id-unrelated features are completely disentangled.

\subsubsection{Domain-Invariant Feature Disentanglement}\label{section:DIL}
Here the content encoder $E_{1}$ is learned by embedding the camera classifier $W_{2}$ for domain-invariant feature extraction. Theoretically, if features extracted by $E_{1}$ are free from various domains, they cannot be assigned to any camera category. To achieve this purpose, $W_{2}$ is designed to enjoy $(C^{s}+C^{t}+1)$ neurons, so that by forwarding $f\_content_{i}^{s}$ / $f\_content_{j}^{t}$ into $W_{2}$, the output is classified to the $(C^{s}+C^{t}+1)$-th class, being different from assignments in the both source domain and target domain. The correlation between $E_{1}$ and $W_{2}$ can be measured as
\begin{equation}
\begin{aligned}
L_{e1}(E_{1})&=
CE\left(W_{2}\left(f\_content_{i}^{s}\right),p_{c}\right)\\
&+CE\left(W_{2}\left(f\_content_{j}^{t}\right),p_{c}\right)\\
&+CE\left(W_{1}\left(f\_content_{i}^{s}\right),y_{i}^{s}\right)
\end{aligned}\label{eq:Lcam1}
\end{equation}
where $p_{c}$ denote the assignment belonging to the $(C^{s}+C^{t}+1)$-th class. In Eq.(\ref{eq:Lcam1}), the first two terms are used for domain-invariant feature learning, and the third term is used for discriminative identity feature learning. Note that, in this equation, only $E_{1}$ is trained to generate domain-invariant features, while the weights in $W_{2}$ are fixed. Since $W_{2}$ is pretrained to be capable of classifying various cameras, and if features learned from $E_{1}$ are only classified to the $(C^{s}+C^{t}+1)$-th class, then there is a high probability that few camera- or domain-related information is contained in these features.


\subsubsection{Id-Unrelated Feature Disentanglement}\label{section:CSL}

After updating the content encoder $E_{1}$, we further exploit the ID classifier $W_{1}$ to handle the domain encoder $E_{2}$, which is implemented through
\begin{equation}
\begin{aligned}
L_{e2}(E_{2})&=
CE\left(W_{2}\left(E_{2}(\mathbf{x}_{i}^{s})\right),c_{i}^{s}\right)+CE\left(W_{2}\left(E_{2}(\mathbf{x}_{j}^{t})\right),c_{j}^{t}\right)\\
&+CE\left(W_{1}\left(E_{2}(\mathbf{x}_{i}^{s})\right),\frac{1}{M^{s}}\right)
\end{aligned}\label{eq:Lid1}
\end{equation}
where $CE\left(W_{1}\left(E_{2}(\mathbf{x}_{i}^{s})\right),\frac{1}{M^{s}}\right)$ means that $W_{1}\left(E_{2}(\mathbf{x}_{i}^{s})\right)$ is enforced to be a $M^{s}\times 1$ vector whose elements are all close to $\frac{1}{M^{s}}$. In other words, the sample $\mathbf{x}_{i}^{s}$ in the source set is averagely classified to all categories, which means unclassified. From Eq.(\ref{eq:Lid1}) we can see that, the first two terms encourage $W_{2}$ to be able to extract the domain-style features, while the third term aim to enforce $W_{1}$ to be unable to classify $E_{2}(\mathbf{x}_{i}^{s})$, proving that there is no any id-related information in these features extracted by $E_{2}$.

\subsubsection{Reciprocal Learning}\label{section:RL}
In the aforementioned analysis, the content encoder $E_{1}$ and domain encoder $E_{2}$ are updated for the domain-invariant features and domain-style features learning based on the camera classifier $W_{2}$ and id classifier $W_{1}$, respectively. Intuitively, $W_{1}$ and $W_{2}$ play key roles in model training. However, both of them are pretrained without any updating, subsequently limiting the performances. To tackle this problem, we introduce an adversarial learning strategy to finetune $W_{1}$ and $W_{2}$, reciprocally contributing to learning the two streams.

Specifically, after updating $E_{1}$ and $E_{2}$, we then update $W_{2}$ and $W_{1}$ through
\begin{equation}\small
\begin{aligned}
L_{adv1}(W_{2})&=
CE\left(W_{2}\left(E_{2}(\mathbf{x}_{i}^{s})\right),c_{i}^{s}\right)
+CE\left(W_{2}\left(E_{2}(\mathbf{x}_{j}^{t})\right),c_{j}^{t}\right)\\
&+CE\left(W_{2}\left(f\_content_{i}^{s}\right),c_{i}^{s}\right)
\\&+CE\left(W_{2}\left(f\_content_{j}^{t}\right),c_{j}^{t}\right)
\end{aligned}\label{eq:Ladv1}
\end{equation}
\begin{equation}\small
\begin{aligned}
L_{adv2}(W_{1})=
CE\left(W_{1}\left(f\_content_{i}^{s}\right),y_{i}^{s}\right)
+CE\left(W_{1}\left(E_{2}(\mathbf{x}_{i}^{s})\right),y_{i}^{s}\right).
\end{aligned}\label{eq:Ladv2}
\end{equation}
From Eq.(\ref{eq:Ladv1}) and Eq.(\ref{eq:Ladv2}), it is easy to observe that $W_{2}$/$W_{1}$ is inversely enforced to classify the id-related/id-unrelated features to their corresponding camera/person IDs. Thanks to these updated classifiers with more powerful classification abilities via the adversarial strategy, we can further improve encoders $E_{1}$ and $E_{2}$ according to Eq.(\ref{eq:Lcam1}) and Eq.(\ref{eq:Lid1}), respectively, greatly contributing to learning domain-invariant features.

\subsection{Optimization}
By jointly taking all loss functions into account, we define the objective function of DRDL as
\begin{equation}
\begin{aligned}
&L(E_{1}, W_{1}, E_{2}, W_{2})=L_{id}(E_{1}, W_{1})+L_{un\_id}(E_{2}, W_{2})+\\ &\alpha L_{e1}(E_{1})+\beta L_{adv1}(W_{2})+\lambda L_{e2}(E_{2})+\tau L_{adv2}(W_{1}).
\end{aligned}\label{eq:objective}
\end{equation}
where $\alpha$, $\beta$, $\lambda$, and $\tau$ are non-negative parameters to tune the importance of different terms. Here we empirically set them to 0.01, 1, 0.01, and 0.01, respectively. In the optimization, we first used $L_{id}(E_{1}, W_{1})$ and $L_{un\_id}(E_{2}, W_{2})$ to pretrain encoders and classifiers. Then according to the adversarial strategy, $\alpha L_{e1}(E_{1})+\beta L_{adv1}(W_{2})+\lambda L_{e2}(E_{2})+\tau L_{adv2}(W_{1})$ focus on learning domain-invariant features and id-unrelated features adversarially.

\begin{figure}
\begin{center}
   \includegraphics[width=1\linewidth]{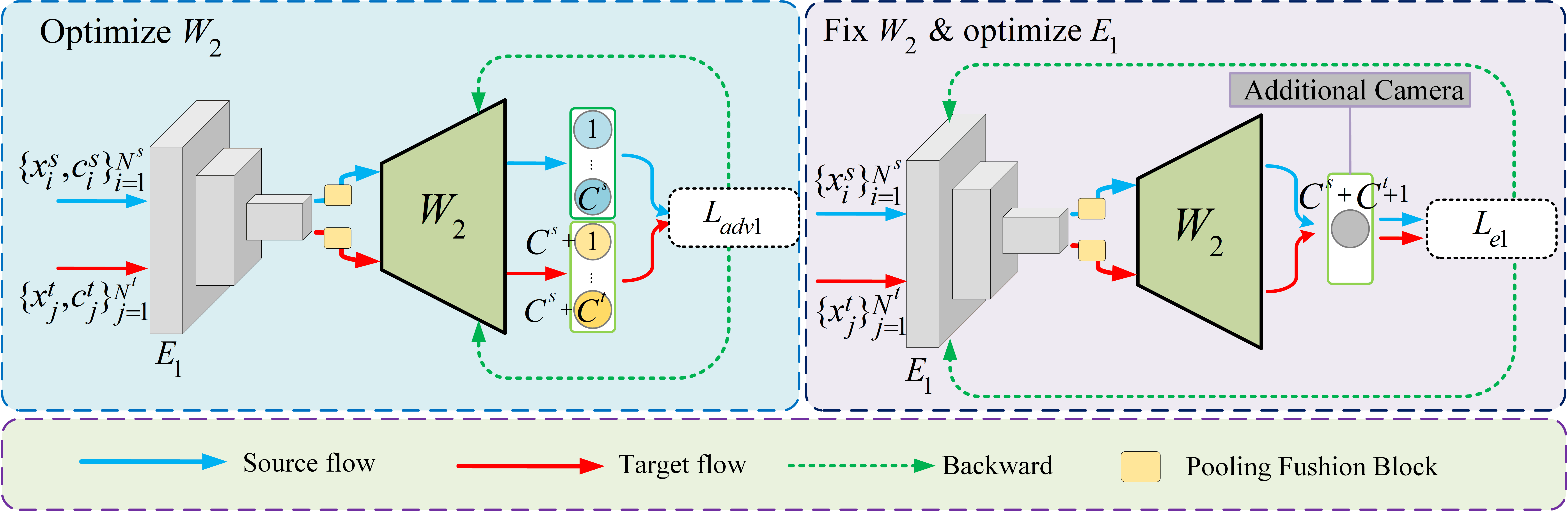}
\end{center}
   \caption{The optimizations for $W_{2}$ and $E_{1}$. By minimizing $L_{adv1}(W_{2})$ with respect to $W_{2}$, $f\_content_{i}^{s}$ / $f\_content_{j}^{t}$ would be classified to a category belonging to $\{1,\cdots, C^{s},(C^{s}+1),\cdots,(C^{s}+C^{t})\}$ through $W_{2}$. Inversely, by minimizing $L_{e1}(E_{1})$, $E_{1}$ is improved to make $W_{2}$ unable to classify $f\_content_{i}^{s}$ / $f\_content_{j}^{t}$ to any camera ID.}
\label{fig:w2update}
\end{figure}

\begin{figure}
\begin{center}
   \includegraphics[width=1\linewidth]{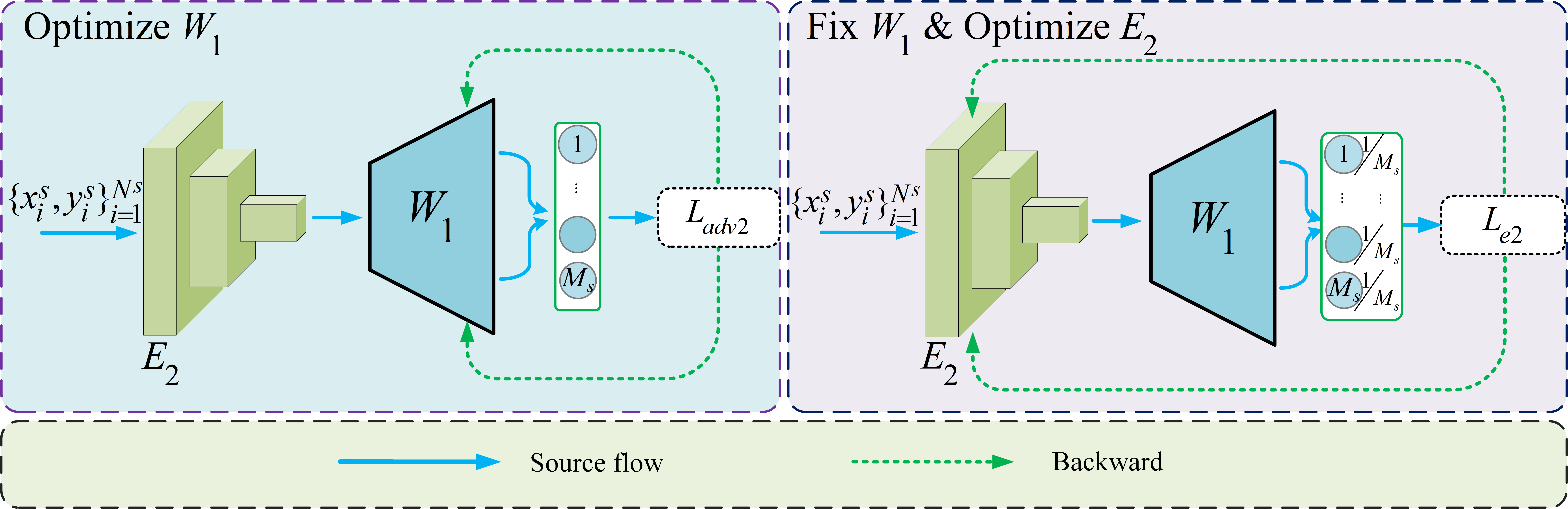}
\end{center}
   \caption{The optimizations for $W_{1}$ and $E_{2}$. By minimizing $L_{adv2}(W_{1})$  with respect to $W_{1}$, $E_{2}(\mathbf{x}_{i}^{s})$ would still be classified to the ID $y_{i}^{s}$. Subsequently, $W_{1}$ is then fixed, and $E_{2}$ is updated to make $W_{1}$ unable to classify $E_{2}(\mathbf{x}_{i}^{s})$ to any person ID via minimizing $L_{e2}(E_{2})$.}
\label{fig:w1update}
\end{figure}

\begin{algorithm}[!tbp]\small
\caption{Dual-Stream Reciprocal Disentanglement Learning}
\begin{algorithmic}[1]
\renewcommand{\algorithmicrequire}{\textbf{Input:}}
\renewcommand{\algorithmicensure}{\textbf{End}}
\REQUIRE samples in the source domain $(\mathbf{x}_{i}^{s}, y_{i}^{s}, c_{i}^{s}) (i = 1,2,\cdots,N^{s})$ and samples in the target domain $(\mathbf{x}_{j}^{t}, c_{j}^{t}) (j = 1,2,\cdots,N^{t})$.
\renewcommand{\algorithmicrequire}{\textbf{Output:}}
\renewcommand{\algorithmicensure}{\textbf{End}}
\REQUIRE The content encoder $E_{1}$.
\renewcommand{\algorithmicrequire}{\textbf{Initialization:}}
\renewcommand{\algorithmicensure}{\textbf{End}}
\REQUIRE Initialize weights of $E_{1}$, $W_{1}$, $E_{2}$ and $W_{2}$, respectively.
\WHILE {not converged or the maximum iteration is not reached}
\IF {$epoch \leqslant iter\_pre$}
\STATE update $E_{1}$ and $W_{1}$ through Eq.(\ref{eq:Lpersonid}),
\STATE update $E_{2}$ and $W_{2}$ through Eq.(\ref{eq:Lcameraid}),
\ELSE
\STATE update $W_{2}$ through Eq.(\ref{eq:Ladv1}),
\STATE update $E_{1}$ through Eq.(\ref{eq:Lcam1}),
\STATE update $W_{1}$ through Eq.(\ref{eq:Ladv2}),
\STATE update $E_{2}$ through Eq.(\ref{eq:Lid1}).
\ENDIF
\ENDWHILE
\end{algorithmic}\label{algorithm:DRDL}
\end{algorithm}

Fig.\ref{fig:w2update} and Fig.\ref{fig:w1update} show alternative optimizations for $W_{2}$/$W_{1}$ and $E_{1}$/$E_{2}$, respectively. In Fig.\ref{fig:w2update}, by minimizing Eq.(\ref{eq:Ladv1}), $W_{2}$ is learned to be capable of assigning id-related features to their associated camera IDs. This means these id-related features also contain more or less id-unrelated information, demonstrating that the content encoder $E_{1}$ is not powerful enough. Therefore, we then fix $W_{2}$ but only optimize Eq.(\ref{eq:Lcam1}) with respect to $E_{1}$, inversely enforcing id-related features to be classified to an additional category through $W_{2}$. In this way $E_{1}$ is further improved to be more powerful in domain-variant feature learning. Similarly, referring to the optimizations for $W_{1}$ and $E_{2}$ in Fig.\ref{fig:w1update}, the content classifier $W_{1}$ is first updated to encourage id-unrelated features $E_{2}(\mathbf{x}_{i}^{s})$ to be classified to its corresponding person ID via Eq.(\ref{eq:Ladv2}). However, by fixing $W_{1}$, $E_{2}$ is then achieved by making $W_{1}$ unable to classify id-unrelated features to their person IDs through Eq.(\ref{eq:Lid1}). Based on these adversarial optimizations, $E_{1}$ is consequently strong enough to extract domain-invariant features for unsupervised person Re-ID. The detailed optimization steps can be found in Algorithm \ref{algorithm:DRDL}.

\begin{table*}\small
\caption{Comparison of the proposed method with state-of-the-art UDA methods on on Duke$\rightarrow$Market and Market$\rightarrow$Duke. The CMC and Map rates (\%) of each method are listed. Duke$\rightarrow$Market means Duke is used as source domain and Market is used as target domain. Vice verse. `--' denotes not reported.}
\centering
\begin{tabular}{|C{1.8cm}|C{1.2cm} C{1.2cm} C{1.2cm} C{1.2cm}|C{1.2cm} C{1.2cm} C{1.2cm} C{1.2cm}|}
\hline
\multirow{3}{*}{Methods}&\multicolumn{4}{c|}{Duke$\rightarrow$ Market}&\multicolumn{4}{c|}{Market$\rightarrow$Duke}\\
\cline{2-9}
{}&{Rank-1}&{Rank-5} &{Rank-10}  &{mAP} &{Rank-1}&{Rank-5} &{Rank-10}  &{mAP}\\
\cline{2-9}
{}&\multicolumn{8}{c|}{Clustering based Methods}\\
\hline
{CAMEL\cite{yu2017cross}}&{54.5}& 73.1& 79.7&{26.3} &{40.3} &57.6 &64.1 &{19.8}\\

{PUL\cite{fan2018unsupervised}}&44.7 &59.1 &65.6 &20.1 &30.4 &44.5 &50.7 &16.4\\

{SSG\cite{fu2019self}}&80.0 &90.0 &92.4 &58.3 &73.0 &80.6 &83.2 &53.4\\

{UDAP\cite{song2020unsupervised}}&75.8 &89.5 &93.2 &53.7 &68.4 &80.1 &83.5 &49.0\\

{ACT\cite{yang2020asymmetric}}&{80.5} &-- &-- &{60.6} &{72.4} &-- &-- &{54.5}\\

{SPCL\cite{ge2020self}}&{90.3} &96.2 &97.7 &{76.7} &{82.9} &90.1 &92.5 &{68.8}\\

{MMT\cite{ge2020mutual}}&87.7 &94.9 &96.9 &71.2 &78.0 &88.8 &92.5 &65.1\\

{DGNet++\cite{zou2020joint}}&82.1 &90.2 &92.7 &61.7 &78.9 &87.8 &90.4 &63.8\\

{MEB-Net\cite{zhai2020multiple}}&89.9 &96.0 &97.5 &76.0 &79.6 &88.3 &92.2 &66.1\\

{DCF\cite{LI2021113}}&86.1 &94.2 &96.0 &67.6 &75.8 &86.5 &89.4 &58.3\\

{IDM\cite{dai2021idm}}&93.2 &97.5 &98.1 &82.8 &83.6 &91.5 &93.7 &70.5\\

\hline
{}&\multicolumn{8}{c|}{Extra Auxiliary Model based Methods}\\
\hline
{PTGAN\cite{wei2018person}}&38.6 &-- &66.1 &-- &27.4 &-- &50.7 &--\\

{ATNet\cite{liu2019adaptive}}&55.7 &73.2 &79.4 &25.6 &45.1 &59.5 &64.2 &24.9\\

{CamStyle\cite{kumar2020unsupervised}}&58.8 &78.2 &84.3 &27.4 &48.4 &62.5 &68.9 &25.1\\

{SBSGAN\cite{huang2019sbsgan}}&58.5 &-- &-- &27.3 &53.5 &-- &-- &30.8\\

{ECN\cite{zhong2019invariance}}&75.1 &87.6 &91.6 &43.0 &63.3 &75.8 &80.4 &40.4\\

{CSGLP\cite{ren2019domain}}&61.2 &77.5 &83.2 &31.5 &47.8 &62.3 &68.3 &27.1\\

{LVRP\cite{yang2019leveraging}}&63.9 &81.1 &86.4 &33.9 &36.3 &54.0 &61.6 &17.9\\
\hline
{}&\multicolumn{8}{c|}{Domain-invariant based Methods}\\
\hline

{CDIL\cite{yuan2020calibrated}}&{57.2} &73.0 &80.0 &{27.4} &{--} &-- &-- &{--}\\

{CFSM\cite{chang2019disjoint}}&{61.2} &-- &-- &{28.3} &{49.8} &-- &-- &{27.3}\\

{CASC\cite{wu2019unsupervised}}&64.7 &80.2 &85.6 &35.6 &51.5 &66.7 &71.7 &30.5\\

{PAUL\cite{yang2019patch}}&{66.7} &-- &-- &{36.8} &{56.1} &-- &-- &{35.7}\\


{FMC\cite{zhang2019fuzzy}}&63.4 &79.5 &84.8 &32.4 &48.0 &62.3 &68.1 &27.8\\

{SSAE\cite{li2020structure}}&{60.7} &-- &-- &{26.6} &{50.2} &-- &-- &{28.1}\\

{SNR\cite{jin2020style}}&{66.7} &-- &-- &{33.9} &{55.1} &-- &-- &{33.6}\\

{MMCL\cite{wang2020unsupervised}}&{66.6} &-- &-- &{35.3} &{58.0} &-- &-- &{36.3}\\

{DGNet++\cite{zou2020joint}}&52.2 &70.7 &77.0 &28.6 &53.2 &68.7 &73.8 &36.3\\

{CAC\cite{li2021cross}}&69.4 &82.8 &87.3 &36.9 &57.5 &71.2 &75.3 &37.0\\


{DRDL}&{{72.7}} &{85.1} &{89.4} &{{38.2}} &{{65.3}} &\textbf{78.7} &\textbf{82.2} &{{42.9}}\\

{DRDL-IBN}&{\textbf{76.8}} &\textbf{88.5} &\textbf{91.6} &{\textbf{42.7}} &{\textbf{65.3}} &{76.9} &{80.7} &{\textbf{43.2}}\\
\hline
\end{tabular}\label{tab:compareD2M}
\end{table*}

Note that, although it seems that $E_{2}$ is only designed for id-unrelated feature extraction which is separative from the domain-invariant feature extraction, it does make an effect on $E_{1}$ learning. As shown in Eq.(\ref{eq:Ladv1}) and Eq.(\ref{eq:Ladv2}), $E_{2}$ are embedded into these two equations for updating $W_{2}$ and $W_{1}$. Obviously, the updated $W_{2}$ and $W_{1}$ will subsequently affect $E_{1}$. Therefore, $E_{2}$, $W_{1}$ and $W_{2}$ are all beneficial for the learning of $E_{1}$.

Furthermore, it seems that Eq.(\ref{eq:Lcameraid}) and Eq.(\ref{eq:Lid1}) have some overlapping. However, we only apply Eq.(\ref{eq:Lcameraid}) to model pretraining, while Eq.(\ref{eq:Lid1}) is exploited for adversarial learning, as shown in Algorithm \ref{algorithm:DRDL}. Thus, they are independent and it is reasonable to preserve the first two terms in Eq.(\ref{eq:Lid1}).

Besides, Eq.(\ref{eq:Lcam1}) enforces $E_{1}$ to extract only id-related features. If the feature obtained by E1 are unrelated to the camera-style, it does not belong to any camera style but an additional one. Thus, we classify it to another category. For Eq.(\ref{eq:Lid1}), it focuses on learning the camera-related feature from $E_{2}$. If we also classify this feature to another category by $W_{1}$, this means images from different cameras enjoy the same identity information. However, this id-related information is much more complex than the camera-related information and using only $E_{2}$ to transform all images to a id-consistent feature is too strict. By contrast, averaging the prediction to each class not only follow our adversarial learning but also is much easier to achieve.

\subsection{Implementation Details}

As we have mentioned above, the ResNet50 is used to be the backbone for two encoders. Besides, the size of each image is resized to $256\times128$. Being similarly to \cite{zhong2020random}, random cropping, horizontal flipping, and random erasing are used for data augmentation. Referring to the matching evaluation, the Euclidean Distance is exploited as the similarity measurement. Furthermore, we implement DRDL by PyTorch \cite{pytorch} with CUDA version 1.6.0, and the model runs on one GPU of NVIDIA  GeForce RTX 2080Ti.

\begin{table*}\small
\caption{Comparison of the proposed method with state-of-the-art UDA methods on on MSMT17$\rightarrow$ Market, MSMT17$\rightarrow$ Duke, Market$\rightarrow$ MSMT17, and Duke$\rightarrow$ MSMT17. The CMC and Map rates (\%) of each method are listed. MSMT17$\rightarrow$ Market means MSMT17 is used as source domain and Market is used as target domain. So do other experiments. `--' denotes not reported.}
\centering
\begin{tabular}{|C{1.6cm}| C{1.2cm} C{1.2cm} C{1.2cm} C{1.2cm}| C{1.2cm} C{1.2cm} C{1.2cm} C{1.2cm}| }
\hline
\multirow{2}{*}{Methods}&\multicolumn{4}{c|}{MSMT17$\rightarrow$ Market}&\multicolumn{4}{c|}{MSMT17$\rightarrow$ Duke}\\
\cline{2-9}
{}&{Rank-1} &{Rank-5} &{Rank-10} &{mAP} &{Rank-1} &{Rank-5} &{Rank-10} &{mAP}\\
\hline
{CASC\cite{wu2019unsupervised}}&{65.4} &80.6 &86.2 &{35.5} &{59.3} &73.2 &77.8 &{37.8} \\


{CDIL\cite{yuan2020calibrated}}&{59.1} &75.4 &-- &{30.3} &{60.7} &74.7 &-- &{39.1}\\

{CAC\cite{li2021cross}}&{72.7} &85.1 &88.8 &{41.0} &{68.0} &80.3 &84.3 &{47.4}\\

DRDL &{\textbf{77.0}} &\textbf{87.6} &\textbf{91.2} &{\textbf{41.7}} &{\textbf{69.9}} &\textbf{82.0} &\textbf{85.6} &{\textbf{48.5}}\\

DRDL-IBN &{\textbf{80.3}} &\textbf{90.9} &\textbf{93.9} &{\textbf{48.7}} &{\textbf{71.9}} &\textbf{84.0} &\textbf{87.8} &{\textbf{51.5}}\\

\hline
\multirow{2}{*}{Methods}&\multicolumn{4}{c|}{Market$\rightarrow$ MSMT17}&\multicolumn{4}{c|}{Duke$\rightarrow$ MSMT17}\\
\cline{2-9}
{}&{Rank-1} &{Rank-5} &{Rank-10} &{mAP} &{Rank-1} &{Rank-5} &{Rank-10} &{mAP}\\
\hline
{PTGAN\cite{wei2018person}}&10.2 &-- &24.4 &2.9 &11.8 &-- &27.4 &3.3\\

{ECN\cite{zhong2019invariance}}&25.3 &36.3 &42.1 &8.5 &30.2 &41.5 &46.8 &10.2\\

{CAC\cite{li2021cross}}&{29.3} &40.2 &45.9 &{10.5} &{37.0} &49.9 &55.6 &{13.3}\\

DRDL & {\textbf{33.8}} &\textbf{45.5} &\textbf{51.2} &{\textbf{12.3}} &{\textbf{41.4}} &\textbf{53.4} &\textbf{58.8} &{\textbf{14.8}}\\

DRDL-IBN &{\textbf{38.6}} &\textbf{51.4} &\textbf{57.1} &{\textbf{14.7}} &{\textbf{42.0}} &\textbf{53.7} &\textbf{59.1} &{\textbf{14.9}}\\
\hline
\end{tabular}\label{tab:comparison2}
\end{table*}

In the training phase, the maximum of epochs is set to 160, in which the first 70 epochs(i.e. the $iter\_pre$ in Algorithm \ref{algorithm:DRDL}) are exploited for pretraining by minimizing Eq.(\ref{eq:Lpersonid}) and Eq.(\ref{eq:Lcameraid}). Beginning from the 71-st epoch, encoders and classifiers are further updated by following Fig.\ref{fig:w2update} and Fig.\ref{fig:w1update}. For the content encoder $E_{1}$, its learning rate is initialized to $3\times 10^{-4}$ with the weight decay $5\times 10^{-4}$. According to \cite{luo2019bag}, we adopt a warm-up technique to linearly tune the learning rate for different epochs. Additionally, referring to $E_{2}$, $W_{1}$, and $W_{2}$, we empirically fix their learning rates to $3.5\times 10^{-4}$, $3\times 10^{-4}$, and $3\times 10^{-5}$, respectively. For all networks, the batchsize is set to 16 and they are optimized through the Adam optimizer.

\section{Experiments}\label{section:experiments}
To show the effectiveness of DRDL, in this sections, experiments with two settings are conducted. Firstly, the datasets which are widely used in existing methods are conducted. To further show the superiority of our proposed method on the case that a person is captured by only one or few of cameras, we then re-conduct the datasets so that parts of identities in them only appear in one camera. Finally, we make ablation study and parameter analysis to analyze the insights of DRDL.

Referring to the evaluation protocols, in our experiments, the Cumulative Matching Characteristics(CMC) curve \cite{wang2007shape} and the mean Average Precision (mAP) \cite{zheng2015scalable} are utilized for all experiment evaluations.

\subsection{Datasets And Compared Methods}
Here three widely used datasets Market1501 (Market) \cite{zheng2015scalable}, DukeMTMC-reID (Duke) \cite{ristani2016performance}, MSMT17 \cite{zheng2017unlabeled}, as well as two more challenged datasets GRID \cite{loy2010time} and PRID2011 \cite{hirzer2011person}, are exploited. \textbf{Market} consists of 6 cameras and 1501 identities, where 751 identities including 12,936 images are used for training and the remaining 750 identities including 19,732 images are used for testing. \textbf{Duke} is composed of 36,411 images associated with 1,812 persons, in which the half number of persons including 16,522 images are used for training and the rest of identities including 17,661 images are used for testing. \textbf{MSMT17} is a much larger set collected from 15 cameras. There are totally 4,101 identities and 126,441 images. Here 32,621 images corresponding to 1,041 identities are used for training and the remaining images are used for testing. \textbf{GRID} was captured from 8 cameras in the railway station, containing 250 pairs of pedestrian images. Each pair was collected from two different cameras. To be more practical, we randomly select 125 pedestrians as well as 400 outlier-images for training, and the rest of 125 pairs corrupted with 375 additional outlier-images are regarded as the test set. \textbf{PRID2011} was collected from two cameras: A and B. Specifically, 385 pedestrians were recorded by Camera A and 749 pedestrians were recorded by Camera B, among which there are 200 overlapped identities. In this paper, 100 identities including 200 images from the overlapped subset are used for training, which are also corrupted with 300 outlier-images. We then respectively select 100 pedestrians collected by Camera A and Camera B as the probe set and the gallery set. Note that the gallery set is also corrupted with 249 outlier-images.

To quantitative evaluate the superiority of our proposed method, aforementioned three types unsupervised Re-ID methods are used for comparison, including (a) clustering based methods, including CAMEL\cite{yu2017cross}, PUL\cite{fan2018unsupervised}, SSG\cite{fu2019self}, UDAP\cite{song2020unsupervised}, ACT\cite{yang2020asymmetric}, SPCL\cite{ge2020self}, MMT \cite{ge2020mutual}, DGNet++\cite{zou2020joint}, MEB-Net\cite{zhai2020multiple}, DCF\cite{LI2021113}, and IDM\cite{dai2021idm}; (b) extra auxiliary model based methods, including PTGAN\cite{wei2018person}, ATNet\cite{liu2019adaptive}, CamStyle\cite{kumar2020unsupervised}, SBSGAN\cite{huang2019sbsgan}, ECN\cite{zhong2019invariance}, CSGLP\cite{ren2019domain}, and LVRP\cite{yang2019leveraging}; (c) domain-invariant learning based methods, including CDIL\cite{yuan2020calibrated}, CFSM\cite{chang2019disjoint}, CASC\cite{wu2019unsupervised}, PAUL\cite{yang2019patch}, FMC\cite{zhang2019fuzzy}, SSAE\cite{li2020structure}, SNR\cite{jin2020style}, MMCL\cite{wang2020unsupervised}, DGNet++\cite{zou2020joint} (without self learning), and CAC\cite{li2021cross}.

\subsection{Experiments Under General Settings}



The experimental results on Duke$\rightarrow$Market and Market$\rightarrow$Duke are listed in Tab.\ref{tab:compareD2M}. It is easy to observe that, although our proposed method DRDL is inferior to some clustering based approaches, it achieves much better performances compared with existing domain-invariant learning based methods. Specifically, DRDL-IBN gains as high as 76.8\%/42.7\% and 65.3\%/43.2\% on Rank-1/mAP, respectively. However, the best performances obtained by CAC among these types of approaches are only 69.4\%/36.9\% and 57.5\%/37.0\%, being inferior to ours'. In contrast to the extra auxiliary model based methods, DRDL also obtains competitive results. Compared with PTGAN, ATNet, CamStyle, SBSGAN, CSGLP and LVRP, our proposed method is much more outstanding. Although ECN is similar to DRDL on Duke$\rightarrow$Market, it still meets an inferior performance on Market$\rightarrow$Duke.

Tab.\ref{tab:comparison2} further displays results on MSMT17$\rightarrow$ Market, MSMT17$\rightarrow$ Duke, Market$\rightarrow$ MSMT17, and Duke$\rightarrow$ MSMT17, compared with existing benchmarks. It is easy to observe that, in these experiments, our proposed method outperforms the most of state-of-the-arts.

In Tab.\ref{tab:compareD2M}, the extra auxiliary model based method ECN gains competitive results. However, due to the image generation, it takes large time costs for training. Tab.\ref{tab:time} makes a comparison of time costs between ECN and DRDL. As we can see, ECN takes approximate 90 hours on Market$\rightarrow$ Duke, while our proposed method only costs about \textcolor{blue}{17} hours, being much more efficient. Furthermore, given a larger dataset, ECN remarkably increases the time cost. For instance, transferring Market to MSMT17, the training time is risen from 90 hours to 420 hours. By contrast, DRDL only takes 21 hours, further showing that DRDL is more adaptive for practical applications.
\subsection{Experiments Under More Practical Settings}\label{section:practical}

\begin{table}\small
\caption{Time costs and Rank-1 obtained by ECN and DRDL on Market$\rightarrow$Duke and Market$\rightarrow$ MSMT17, respectively.}
\centering
\begin{tabular}{|C{1.2cm}| C{1.1cm} C{1.1cm}| C{1.1cm} C{1.1cm}| }
\hline
\multirow{2}{*}{Methods}&\multicolumn{2}{c|}{Market$\rightarrow$Duke} &\multicolumn{2}{c|}{Market$\rightarrow$MSMT17}\\
\cline{2-5}
{}&{Rank-1} &{Times/h} &{Rank-1} &{Times/h}\\
\hline
{ECN\cite{zhong2019invariance}}&{63.3} &{$\approx$ 90} &{25.3} &{$\approx$ 420}\\

{DRDL}&{\textbf{65.3}} &{$\approx$ \textbf{17}} &{\textbf{33.8}} &{$\approx$\textbf{21}}\\
\hline
\end{tabular}\label{tab:time}
\end{table}

\begin{table*}\small
\caption{Comparison of the proposed method with state-of-the-art UDA methods on on Market$\rightarrow$PRID2011, Market$\rightarrow$ GRID, Duke$\rightarrow$ PRID2011, and Duke$\rightarrow$ GRID. The CMC and Map rates (\%) of each method are listed. Market$\rightarrow$PRID2011 means Market is used as source domain and PRID2011 is used as target domain. So do other experiments.}
\centering
\begin{tabular}{|C{2cm}| C{1.2cm} C{1.2cm} C{1.2cm} C{1.2cm}| C{1.2cm} C{1.2cm} C{1.2cm} C{1.2cm}|}
\hline
\multirow{2}{*}{Methods}&\multicolumn{4}{c|}{Market$\rightarrow$ PRID2011}&\multicolumn{4}{c|}{Market$\rightarrow$ GRID}\\
\cline{2-9}
{}&{Rank-1} &{Rank-5} &{Rank-10} &{mAP} &{Rank-1} &{Rank-5} &{Rank-10} &{mAP}\\
\hline
{ACT\cite{yang2020asymmetric}}&14.0 &26.0 &38.0 &22.2 &14.4 &31.2 &42.2 &23.4\\

{MMT\cite{ge2020mutual}}&{31.0} &48.0 &57.0 &{39.8} &{32.0} &48.0 &57.0 &{40.3}\\

{UDAP\cite{song2020unsupervised}}&12.0 &23.0 &35.0 &19.5 &28.0 &50.4 &59.2 &38.8\\

{SPCL\cite{ge2020self}}&4.0 &10.5 &16.5 &8.6 &13.6 &24.0 &35.2 &20.2 \\

DRDL &{\textbf{46.2}} &\textbf{71.0} &\textbf{81.4} &{\textbf{67.5}} &{\textbf{46.8}} &\textbf{65.6} &\textbf{72.9} &{\textbf{55.2}}\\

DRDL-IBN & {\textbf{54.9}} &\textbf{75.2} &\textbf{82.4} &{\textbf{64.5}} &{\textbf{47.2}} &\textbf{63.9} &\textbf{70.8} &{\textbf{55.3}}\\

\hline
\multirow{2}{*}{Methods}&\multicolumn{4}{c|}{Duke$\rightarrow$ PRID2011}&\multicolumn{4}{c|}{Duke$\rightarrow$ GRID}\\
\cline{2-9}
{}&{Rank-1} &{Rank-5} &{Rank-10} &{mAP} &{Rank-1} &{Rank-5} &{Rank-10} &{mAP}\\

\hline
{ACT\cite{yang2020asymmetric}}&13.0 &31.0 &40.0 &21.9 &13.6 &25.6 &30.4 &20.4\\

{MMT\cite{ge2020mutual}} &25.0 &41.0 &54.0 &33.9 &32.8 &49.6 &55.2 &41.3\\

{UDAP\cite{song2020unsupervised}} &22.0 &47.0 &55.0 &33.3 &27.2 &44.8 &55.2 &35.6\\

{SPCL\cite{ge2020self}}&16.7 &31.1 &39.6 &7.6 &8.0 &20.8 &33.6 &15.7\\

DRDL &{\textbf{52.4}} &\textbf{76.6} &\textbf{82.4} &{\textbf{63.2}} &{\textbf{39.9}} &\textbf{55.1} &\textbf{64.8} &{\textbf{47.8}}\\

DRDL-IBN &{\textbf{45.8}} &\textbf{63.6} &\textbf{71.2} &{\textbf{54.2}} &{\textbf{34.5}} &\textbf{52.3} &\textbf{60.3} &{\textbf{50.1}}\\
\hline
\end{tabular}\label{tab:robust}
\end{table*}

\begin{figure}
\begin{center}
   \includegraphics[width=1\linewidth]{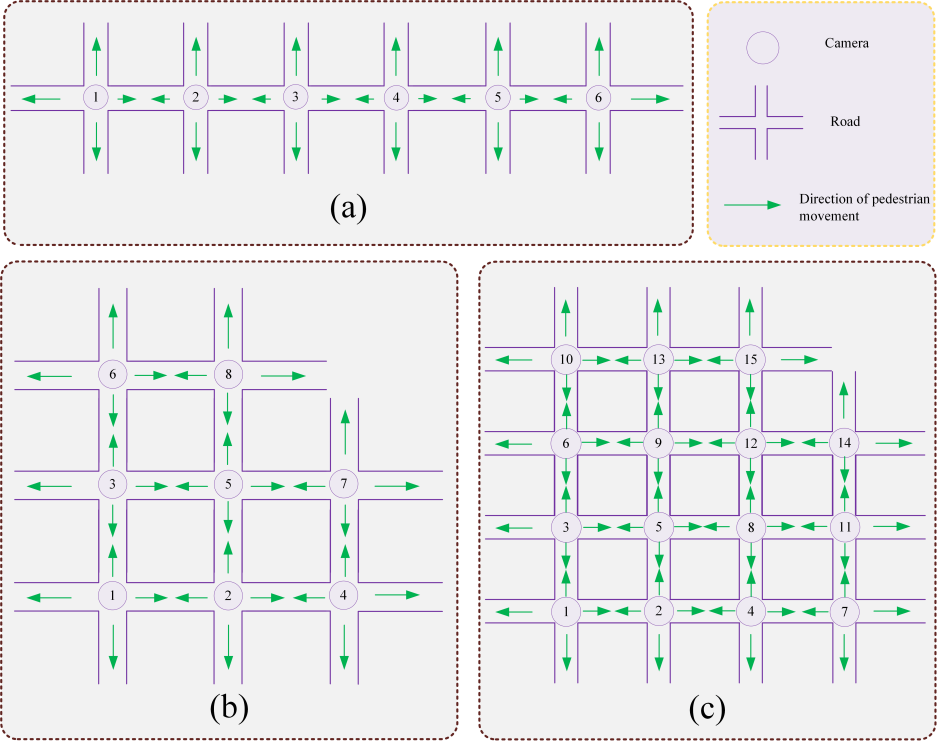}
\end{center}
   \caption{The strategies for Market, Duke and MSMT17 datasets re-conduction, respectively. (a) Market, (b) Duke, (c) MSMT17.}
\label{fig:camera}
\end{figure}

\begin{table*}\small
\caption{The comparison details between the re-conducted datasets and the original versions for Market, Duke, and MSMT17, respectively.}
\centering
\begin{tabular}{|C{2.4cm}| C{1.2cm}| C{1.2cm} C{1.2cm}| C{1.2cm} C{1.2cm} |C{1.2cm} C{1.2cm}| C{1.2cm}|}
\hline
\multirow{2}{*}{Datasets}&\multirow{2}{*}{ID}&\multicolumn{2}{c|}{Train}
&\multicolumn{2}{c|}{Probe(Test)}&\multicolumn{2}{c|}{Gallery(Test)}&\multirow{2}{*}{Cameras}\\
\cline{3-8}
{}&{} &{ID} &{Images}&{ID} &{Images}&{ID} &{Images}&{}\\
\hline
Market&1501&751&12936&750&3368&{750}&19732&6\\

Duke& 1812&702&16522&702&2228&1110&17661&8\\

MSMT17& 4101&1041&32621&3060&11659&3060&{82161}&15\\

Market-new& {1367}&617&3197&750&3368&{750}&{19732}&6\\

Duke-new& {1255}&553&5300&702&2228&1110&17661&8\\

MSMT17-new&{2831}&1790&15356&1041&2900&1041&29721&15\\
\hline
\end{tabular}\label{tab:datasetnew}
\end{table*}

Referring to the clustering based methods, in Tab.\ref{tab:compareD2M}, e.g., SSG, ACT, SPCL, MMT, DG-Net++ and MEB-Net, it is true that they achieve better results in comparison to our presented approach DRDL. However, as we have mentioned above, their outstanding performances are highly dependent on the assumption: each identity appears in enough cameras and have enough positive samples, while it cannot be met in many practical applications. By contrast, our proposed method prevents from estimating the pseudo-labels, being free from this limitation. To make a proof, we further conduct experiments on GRID and PRID2011, which are more challenged with outlier-images and many identities only appear in a camera. Here GRID and PRID2011 are used as target domains, and Market and Duke are exploited as the source domains, respectively. The results are tabled in Tab.\ref{tab:robust}. It is easy to see that, DRDL obtains much higher results, indicating its robustness on real-world applications.

To further demonstrate the effectiveness of our proposed method on the application where some identities were captured by only one or few of cameras, we re-conduct Market, Duke and MSMT17 to follow the requirement. As shown in Fig.\ref{fig:camera}(a), assume that the six cameras in Market are layouted along a road and each person enjoys the same probability to move to four directions (up, down, left and right). Thus, each person has 25\% probability to move from one camera to another linked camera. Specifically, take camera 1 and camera 2 as the example. By following the assumption mentioned above, only 25\% part of identities in camera 1 are allowed to move to camera 2. So is the case from camera 2 to camera 1. Finally, in the training set, 617 identities with 3,197 images are selected. Fig.\ref{fig:camera}(b) displays the camera locations for Duke. Being similar to Fig.\ref{fig:camera}(a), cameras 1, 2, 3, 4, 6, 7, 8 enjoy the possibilities of moving to their linked cameras. Therefore, in the training set, 553 identities with 5300 images are selected. For MSMT17, since the testing set contains more identities than that in the training set, we inversely regard the original testing set as the training set, while the original training set is regarded as the testing set. By following the camera layout in Fig.\ref{fig:camera}(c), 1,790 identities with 15,356 images are re-exploited for training. For the testing set, there are 1,041 identities with 32,621 images, in which 2,900 images are used as the probe set and 29,721 images are used as the gallery set. Here we denote these re-conducted datasets as Market-new, Duke-new, and MSMT17-new, respectively. The comparison details between these re-conducted datasets and the original versions are listed in Tab.\ref{tab:datasetnew}.

\begin{table*}\small
\caption{Comparison of the proposed method with state-of-the-art UDA methods on on Duke$\rightarrow$Market-new, Market$\rightarrow$Duke-new, Duke$\rightarrow$MSMT17-new, and Market$\rightarrow$MSMT17-new. The CMC and Map rates (\%) of each method are listed. Duke$\rightarrow$Market-new means Duke is used as source domain and Market-new is used as target domain. So do other experiments. `--' denotes not reported.}
\centering
\begin{tabular}{|C{2.2cm}|C{1.2cm} C{1.2cm} C{1.2cm} C{1.2cm}|C{1.2cm} C{1.2cm} C{1.2cm} C{1.2cm}|}
\hline
\multirow{3}{*}{Methods}&\multicolumn{4}{c|}{Duke$\rightarrow$ Market-new}&\multicolumn{4}{c|}{Market$\rightarrow$Duke-new}\\
\cline{2-9}
{}&{Rank-1} &{Rank-5} &{Rank-10} &{mAP} &{Rank-1} &{Rank-5} &{Rank-10} &{mAP}\\
\hline

{UDAP\cite{song2020unsupervised}}&56.6 &72.2 &77.5 &30.6 &42.9 &57.2 &63.2 &27.7\\

{ACT\cite{yang2020asymmetric}}&51.5 &67.0 &72.5 &26.0 &30.3 &42.3 &48.9 &18.0\\

{MEB-Net\cite{zhai2020multiple}}&57.3 &73.0 &79.1 &33.4 &44.2 &59.1 &65.4 &30.7\\

{SPCL\cite{ge2020self}}&14.1 &26.1 &33.0 &5.6 &13.2 &21.5 &25.3 &5.5\\

{MMT-500\cite{ge2020mutual}}&59.7 &75.1 &80.9 &33.7 &45.4 &61.0 &67.6 &30.8\\

{MMT-700\cite{ge2020mutual}}&57.3 &73.5 &80.2 &32.0 &45.2 &60.5 &67.0 &30.5\\

{MMT-900\cite{ge2020mutual}}&58.4 &74.3 &80.3 &32.5 &43.9 &61.0 &67.1 &30.8\\

{CAC\cite{li2021cross}}&58.9 &75.0 &80.2 &28.2 &50.5 &65.0 &70.4 &30.7\\

{IDM\cite{dai2021idm}}&42.7 &59.4 &66.9 &20.4 &37.4 &50.4 &57.9 &24.5\\


{DRDL}&\textbf{70.4} &\textbf{84.3} &\textbf{89.0} &\textbf{37.1} &\textbf{62.9} &\textbf{76.4} &\textbf{80.9} &\textbf{41.0}\\
\hline
\hline
\multirow{3}{*}{Methods}&\multicolumn{4}{c|}{Duke$\rightarrow$ MSMT17-new}&\multicolumn{4}{c|}{Market$\rightarrow$MSMT17-new}\\
\cline{2-9}
{}&{Rank-1} &{Rank-5} &{Rank-10} &{mAP} &{Rank-1} &{Rank-5} &{Rank-10} &{mAP}\\
\hline

{UDAP\cite{song2020unsupervised}}&23.3 &35.1 &41.0 &9.8 &14.5 &25.0 &31.2 &6.0\\

{ACT\cite{yang2020asymmetric}}&15.2 &24.2 &29.4 &6.4 &9.5 &18.3 &24.1 &4.0\\

{MEB-Net\cite{zhai2020multiple}}&33.9 &46.5 &52.2 &15.9 &26.1 &37.3 &43.5 &12.0\\

{SPCL\cite{ge2020self}}&19.8 &31.7 &37.7 &8.8 &18.8 &30.4 &36.6 &9.0\\

{MMT-500\cite{ge2020mutual}}&35.8 &47.6 &53.4 &15.3 &27.4 &39.7 &45.7 &11.6\\

{MMT-1000\cite{ge2020mutual}}&36.5 &50.4 &55.9 &16.3 &30.7 &43.9 &50.9 &13.7 \\

{MMT-1500\cite{ge2020mutual}}&39.0 &51.0 &57.8 &17.3 &33.2 &45.2 &51.9 &14.5\\

{MMT-2000\cite{ge2020mutual}}&37.8 &51.7 &57.5 &17.2 &33.9 &47.5 &54.9 &15.4\\

{CAC\cite{li2021cross}}&33.5 &44.1 &50.1 &12.4 &20.7 &32.7 &38.1 &7.6 \\

{IDM\cite{dai2021idm}}&23.2 &33.2 &39.3 &10.2 &20.3 &30.3 &36.2 &8.9\\


{DRDL}&\textbf{48.2} &\textbf{60.0} &\textbf{64.3} &\textbf{19.5} &\textbf{39.0} &\textbf{51.7} &\textbf{56.7} &\textbf{16.2}\\
\hline
\end{tabular}\label{tab:comparenew}
\end{table*}

\begin{table*}\small
\caption{Experimental results obtained by our proposed method under different constraints when id-related features are used to match pedestrian identities. The CMC and MAP rates(\%) of each method are listed. $L_{BL}$ (Baseline) denotes $L_{id}$+$L_{un\_id}$, $L_{1}$ denotes $L_{e1}$ + $L_{adv1}$, and $L_{2}$ denotes $L_{e2}$ + $L_{adv2}$.}
\centering
\begin{tabular}{|C{2.4cm}| C{1.2cm} C{1.2cm} C{1.2cm} C{1.2cm}| C{1.2cm} C{1.2cm} C{1.2cm} C{1.2cm}| }
\hline
\multirow{2}{*}{Methods}&\multicolumn{4}{c|}{Duke$\rightarrow$ Market}&\multicolumn{4}{c|}{Market$\rightarrow$ Duke}\\
\cline{2-9}
{}&{Rank-1} &Rank-5 &Rank-10 &{mAP} &{Rank-1} &Rank-5 &Rank-10 &{mAP}\\
\hline
{BL ($L_{BL}$)}&{64.6} &79.1 &84.2 &{34.4} &{48.8} &64.1 &69.7 &{31.8}\\

{BL+$L_{1}$}&{67.9} &82.3 &87.1 &{35.8} &{57.8} &71.8 &75.8 &{35.5}\\

{BL+$L_{1}$+$L_{2}$}&{68.4} &83.0 &87.8 &{36.3} &{60.0} &74.5 &78.9 &{38.3}\\

{BL+$L_{1}$+$L_{2}$+PFB}&{\textbf{72.7}} &\textbf{85.1} &\textbf{89.4} &{\textbf{38.2}} &{\textbf{65.3}} &\textbf{78.7} &\textbf{82.2} &{\textbf{42.9}}\\
\hline
\end{tabular}\label{tab:ablation}
\end{table*}

\begin{table*}\small
\caption{Experimental results obtained by our proposed method under different constraints when id-unrelated features are used to match pedestrian identities. The CMC and MAP rates(\%) of each method are listed. $L_{BL}$ (Baseline) denotes $L_{id}$+$L_{un\_id}$, $L_{1}$ denotes $L_{e1}$ + $L_{adv1}$, and $L_{2}$ denotes $L_{e2}$ + $L_{adv2}$.}
\centering
\begin{tabular}{|C{2.4cm}| C{1.2cm} C{1.2cm} C{1.2cm} C{1.2cm}| C{1.2cm} C{1.2cm} C{1.2cm} C{1.2cm}| }
\hline
\multirow{2}{*}{Methods}&\multicolumn{4}{c|}{Duke$\rightarrow$ Market}&\multicolumn{4}{c|}{Market$\rightarrow$ Duke}\\
\cline{2-9}
{}&{Rank-1} &Rank-5 &Rank-10 &{mAP} &{Rank-1} &Rank-5 &Rank-10 &{mAP}\\
\hline
{BL ($L_{BL}$)}&{0.5362} &1.8170 &2.8597 &{0.3152} &{0.6739} &1.9317 &3.1447 &{0.3797}\\

{BL+$L_{1}$}&{0.4766} &1.4596 &2.3235 &{0.2618} &{0.0449} &0.2246 &0.2246 &{0.1117}\\

{BL+$L_{1}$+$L_{2}$}&{0.3872} &1.4000 &1.8170 &{0.2400} &{0.0000} &0.0449 &0.1348 &{0.0909}\\

\hline
\end{tabular}\label{tab:ablation_id_ir}
\end{table*}

The experimental results on Market-new, Duke-new, and MSMT17-new are displayed in Tab.\ref{tab:comparenew}. Note that, these re-conducted datasets are only used as target domains in the training stages and all comparison methods follow their original experimental settings. From Tab.\ref{tab:comparenew} we can see that, when the datasets in the target domains are replaced with the re-conducted ones, almost all of approaches meet a performance drop, due to the smaller number of training images. Furthermore, compared with the domain invariant based methods, the clustering based methods, e.g., UDAP, ACT, MEB-Net, SPCL and MMT, meet much larger degradation. Particularly, in the experiment of Duke$\rightarrow$ Market-new, the most three superior methods MEB-Net, SPCL and IDM in Tab.\ref{tab:compareD2M} heavily have more than 32\% decrease. Referring to UDAP and ACT, the Rank-1/mAP metrics also remarkably fall from 75.8\%/53.7\% and 80.5\%/60.6\% to 56.6\%/30.6\% and 51.5\%/26.0\%, respectively. The possible reason is that these approaches are too heavily dependent on the assumption that each identity contains enough positive images in different cameras. However, when a more practical and general dataset is given, in which most of identities are captured by only one or few cameras, the clustering strategies based on this assumption are fail. By contrast, the domain invariant based methods, which are free from the aforementioned constraint, are relatively more robust. Specifically, the Rank-1/mAP results computed by our presented strategy DRDL are only changed from 72.7\%/38.2\% to 70.4\%/37.1\%, which is quite adaptive for the real-world applications. Similarly, experiments on Market$\rightarrow$Duke-new, Duke$\rightarrow$MSMT17-new, and Market$\rightarrow$MSMT17-new also substantiate the effectiveness of our proposed method, when the datasets are more practical and challenge.
\subsection{Ablation Study}
To analyze the insight of the proposed method, in this subsection the ablation study is conducted. Specifically, we regard the network guided by $L_{id}(E_{1},W_{1})$ and $L_{un\_id}(E_{2},W_{2})$ as the `Baseline' (BL). Based on this Baseline, we gradually add domain-invariant feature disentanglement, id-unrelated disentanglement, and Fusion Block into it to prove that all of them contribute to the performance enhancement. Note that all experiments are conducted on Duke and Market in this subsection and GAP is adopted for $E_{1}$ in `Baseline'.
\begin{figure}
\begin{center}
   \includegraphics[width=1\linewidth]{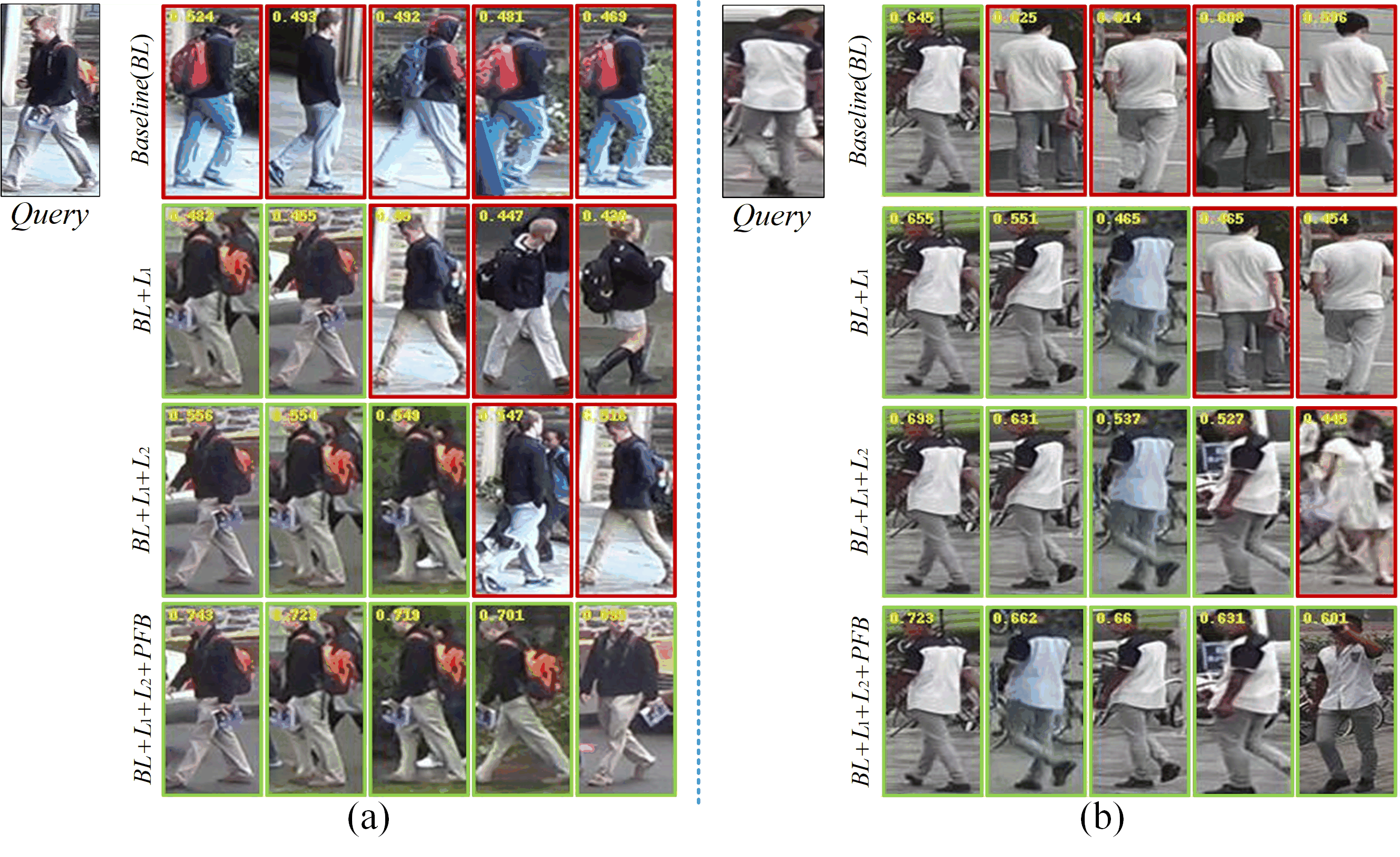}
\end{center}
\caption{Visualization of ablation study. (a) The retrieved results of different Re-ID models on the task of  Market$\rightarrow$Duke. (b)The retrieved results of different Re-ID models on the task of Duke$\rightarrow$Market. The query images and the top-5 retrieved samples are showed. From top to bottom, the results are retrieved by Baseline, BL+$L_{1}$, BL+$L_{1}$+$L_{2}$ and BL+$L_{1}$+$L_{2}$+PFB. Green and red box indicate correct and incorrect retrieved results, respectively.}
\label{fig:visualization}
\end{figure}

\textbf{Domain-Invariant Feature Disentanglement:} To remove the gap between the source domain and target domain, the content encoder $E_{1}$ and the camera classifier $W_{2}$ are learned by following the adversarial strategy, so that $E_{1}$ is capable of extracting domain-invariant features. As listed in Tab.\ref{tab:ablation}, Rank-1/mAP obtained by the Baseline model are only 64.6\%/34.4\% and 48.8\%/31.8\% on the Duke $\rightarrow$ Market and Market $\rightarrow$ Duke, respectively. By contrast, when the domain-invariant feature disentanglement parts including $L_{e1}$ and $L_{adv1}$ are added into, there is an obvious performance improvement, indicating their significance.

\textbf{ID-Unrelated Feature Disentanglement:} The domain-style encoder $E_{2}$ and the identity classifier $W_{1}$ are also embedded for domain-style feature disentanglement (Baseline + $L_{e1}$ + $L_{adv1}$ + $L_{e2}$ + $L_{adv2}$), which are then beneficial for learning $E_{1}$. Experimentally, from Tab.\ref{tab:ablation} we can observe that, by additionally minimizing $L_{adv2}$ and $L_{e2}$, the metrics of rank-1/mAP increase from 67.9\%/35.8\% and 57.8\%/35.5\% to 68.4\%/36.3\% and 60.0\%/38.3\% on the two experiments, respectively, also proving the significance of domain-style feature disentanglement.

\textbf{Pooling Fusion Block:} As mentioned above, to enjoy the merits of GAP and GMP, they are fused to be the pooling fusion block (PFB) in $E_{1}$. Here we further add PFB into (Baseline + $L_{e1}$ + $L_{adv1}$ + $L_{e2}$ + $L_{adv2}$), which meets an increase on both experiments. Thus, using PFB is more powerful to extract the discriminative features.

Additionally, we further display some visualizations in Fig.\ref{fig:visualization}. Obviously, with the embedding of our proposed techniques, our model is more powerful in retrieving, substantiating their significance

\textbf{Retrieval using ID-Unrelated Feature:} To further show the effectiveness of our proposed techniques, we also use id-unrelated features to match pedestrian identities. Theoretically, the less id-related information remaining in identity-unrelated information, the lower the matching rate should be. As listed in Tab.\ref{tab:ablation_id_ir}, when $L_{id}$ and $L_{unid}$ are used for training, Rank-1 on Duke$\rightarrow$Market and Market$\rightarrow$Duke are as high as 0.5362\% and 0.6739\%, respectively. However, with the embedding of $L_{e1}+L_{adv1}$ and $L_{e2}+L_{adv2}$, this metric heavily decreases to 0.3872\% and 0\%, substantiating that our proposed disentanglement learning does encourage the id-unrelated feature to be less discriminative.
\begin{figure*}
\begin{center}
   \includegraphics[width=1\linewidth]{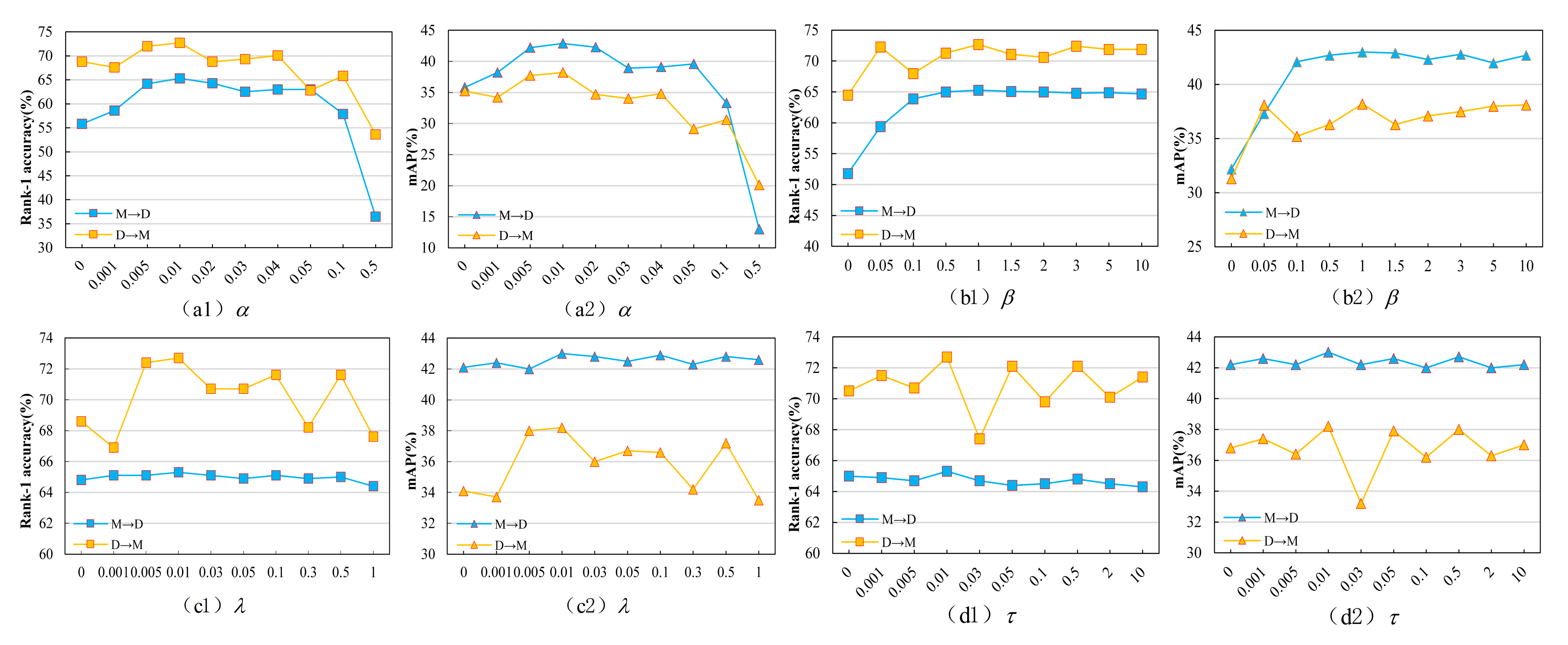}
\end{center}
   \caption{The results of Rank-1 and mAP under the changes of different parameters. M$\rightarrow$D denotes Market$\rightarrow$Duke and D$\rightarrow$M denotes Duke$\rightarrow$Market. }
\label{fig:parameter}
\end{figure*}

\begin{table}\small
\caption{Comparisons of computational costs of different methods on Market$\rightarrow$Duke.}
\centering
\begin{tabular}{|C{2cm}| C{1.3cm} C{1.3cm} C{1.6cm}|}
\hline
{Methods}&{Params/M} &{FLOPs/G} &{Train time/h}\\
\hline
{Baseline}&{25.05} &{4.08} &{$\approx$5}\\
{DRDL}&{56.95} &{8.15} &{$\approx$17}\\
{DG-Net\cite{zheng2019joint}}&{120.95} &{26.49} &{$\approx$35}\\
{DG-Net++\cite{zou2020joint}}&{234.19} &{53.98} &{$\approx$102}\\
\hline
\end{tabular}\label{tab:fuzadu}
\end{table}
\subsection{Training Cost Analysis}
Compared with the most of existing disentanglement based methods, our presented approach is free from image reconstruction, contributing to reducing the computational cost. Quantitatively, the comparison among Baseline, DRDL, DG-Net \cite{zheng2019joint}, and DG-Net++ \cite{zou2020joint} on the training cost is shown in Tab.\ref{tab:fuzadu}. Note that, under the same experimental platform and environment, we run the codes of DG-Net and DG-Net++ again on the task of Market$\rightarrow$ Duke. In this process, we do not make any changes to the code and model hyperparameters, and directly use the parameter settings recommended by the authors. Due to two branches, it is true that metrics on Parameters and Flops are two times compared with that of baseline. However, in contrast to other disentanglement based methods, e.g., DG-Net and DG-Net++, DRDL is also much more efficient. As we can see, parameters and FLOPs of DRDL are only 56.95M and 8.15G, while these of DG-Net/DG-Net++ reach as high as 120.95M/234.19M and 26.49G/53.98G, which are much larger. For the training time, these two comparison methods cost about 35 and 102 hours respectively, which are about two and six times of ours'. Thus, our presented method is much simpler and more efficient.


\subsection{Parameter Selection and Analysis}
In the objective function, four parameters $\alpha$, $\beta$, $\lambda$ and $\tau$ are predefined to tune the significance for different terms. In this subsection, we tune a parameter by manual search with remaining parameters fixed. Also, all experiments are conducted on Duke and Market. Fig.\ref{fig:parameter} displays the experimental results under the changes of different parameters. As we can see, for all parameters, with the increase of their values, the performances meet the enhancement, demonstrating that their associated terms do make contributions. With their continuous rise, however, they then get the performance degradation. The main reason is that a too large value of a parameter may result in over-fitting. Therefore, we set $\alpha$, $\beta$, $\lambda$ and $\tau$ to 0.01, 1, 0.01, and 0.01, respectively. In fact, when there are many hyperparameters that need to be tuned, it seems that some parameter search algorithms such as AutoHAS \cite{ICLR2021} can be used to automatically determine their values. However, the focus of this paper is not to search for the value of hyperparameters. Here we follow most of related works to get the values of hyperparameters empirically.

\section{Conclusion}\label{section:conclusion}
In this paper, a novel domain-invariant feature learning method is proposed for unsupervised person Re-ID. Particularly, two streams are constructed for id-related feature extraction and id-unrelated feature extraction through their corresponding encoders and classifiers. By introducing an adversarial updating strategy, these two streams are reciprocally beneficial for each other, so that id-related and id-unrelated features are disentangled, greatly contributing to learning domain-invariant but discriminative features. The experimental results show that our proposed method not only achieves outstanding Re-ID performances in comparison with previous state-of-the-art methods, but also is efficient on time costs due to the unconstraint on image generation.

\ifCLASSOPTIONcaptionsoff
  \newpage
\fi
\bibliographystyle{IEEEtran}
\bibliography{IEEEfull,egbib}
\end{document}